\documentclass[pdflatex,sn-vancouver-num]{sn-jnl}

\usepackage[utf8]{inputenc}
\usepackage{graphicx}
\usepackage{xcolor}
\usepackage{booktabs}
\usepackage{multirow}
\usepackage{threeparttable}
\usepackage{subcaption}
\usepackage{amsmath,amssymb,amsfonts}
\usepackage{mathrsfs}
\usepackage{algorithm}
\usepackage{algorithmicx}
\usepackage{algpseudocode}
\usepackage{listings}

\usepackage{siunitx}
\begin{document}
\title[Article Title]{From Memorization to Generalization: Fine-Tuning Large Language Models for Biomedical Term-to-Identifier Normalization}

\author[1]{\fnm{Suswitha} \sur{Pericharla}\email{sp5526s@missouristate.edu}}

\author[2,3]{\fnm{Daniel B.} \sur{Hier}\email{dhier@uic.edu}}

\author[3]{\fnm{Tayo} \sur{Obafemi-Ajayi}\email{tayoobafemiajayi@missouristate.edu}}

\affil[1]{\orgdiv{ Computer Science Department}, \orgname{Missouri State University}, \orgaddress{\street{901 S. National Avenue}, \city{Springfield MO}, \postcode{65897}, \state{MO}, \country{USA}}}

\affil[2]{\orgdiv{Department of Neurology and Rehabilitation}, 
\orgname{University of Illinois at Chicago}, 
\orgaddress{\street{912 S. Wood Street}, 
\city{Chicago}, 
\postcode{60612}, 
\state{IL}, 
\country{USA}}}

\affil[3]{\orgdiv{ Engineering Program}, \orgname{Missouri State University}, \orgaddress{\street{901 S. National Avenue}, \city{Springfield MO}, \postcode{65897}, \state{MO}, \country{USA}}}

\abstract{\textbf{Background:}
Effective biomedical data integration depends on automated term normalization—the mapping of natural language biomedical terms to standardized identifiers across terminologies. This linking of terms to identifiers is a crucial step toward semantic interoperability.
Although large language models (LLMs) show promise on this task, they exhibit heterogeneous performance across different biomedical terminologies.
We evaluated both memorization (performance on training terms) and generalization (performance on unseen validation terms) across three biomedical terminologies: the Human Phenotype Ontology (HPO), Gene Ontology (GO), and protein–gene symbol mappings (GENE).

\textbf{Results:}
Fine-tuning Llama 3.1 8B revealed differences across terminologies. GO mappings achieved substantial memorization gains (up to 77\% improvement for term$\rightarrow$identifier on training terms), whereas fine-tuning yielded minimal gains for HPO.
Generalization to unseen validation terms occurred only for protein–gene (GENE) mappings (13.9\% improvement), while fine-tuning on HPO and GO produced negligible generalization. Baseline performance varied by model scale, with GPT-4o consistently outperforming both Llama variants for all terminologies. Embedding analyses revealed strong semantic alignment between gene symbols and protein names but no alignment between terms and identifiers for GO or HPO, consistent with differences in lexicalization.

\textbf{Conclusions:}
The success of fine-tuning in biomedical term normalization depended on two interacting factors: identifier popularity and identifier lexicalization.
Popularity served as a proxy for the likelihood that an LLM encountered a term–identifier pair during pretraining.
Despite the high incidence of HPO terms in the literature, the low frequency of corresponding HPO identifiers created a bottleneck that limited memorization. Lexicalization—the degree to which identifiers convey semantic information through embeddings—enabled generalization in the GENE model, where gene symbols behave as if they were natural language tokens. By contrast, the arbitrary identifiers of GO and HPO constrained models to rote memorization.
These findings offer a predictive framework for understanding when fine-tuning will enhance factual recall and when it will fail due to low popularity or non-lexicalized identifiers.}

\maketitle

\section*{Introduction}\label{sec1}
The exponential growth of biomedical data has created unprecedented opportunities for advancing healthcare through data mining and machine learning. Realizing this potential requires robust data integration across diverse terminologies and datasets. Large language models (LLMs) have shown promise in addressing these challenges, particularly in clinical text processing and concept extraction. LLMs can process clinical notes and extract medical concepts effectively~\cite{hier2025preprocessing,hier2024highthroughput}, making them valuable tools for converting unstructured text into computable biomedical data.

Generative autoregressive LLMs, such as models from the GPT and Llama series, produce text by predicting the next token in a sequence. Owing to their large parameter counts and broad training corpora, they exhibit emergent capabilities, notably the ability to retrieve factual knowledge. Although designed for token prediction, they are increasingly being used as repositories of information. This retrieval ability likely reflects two complementary mechanisms: probabilistic token associations learned during pretraining and structured semantic representations encoded in vector embeddings. Wang et al. \cite{wang2024re} distinguish between fact-based capabilities (e.g., “What is the HPO identifier for tremor?”) and skill-based capabilities (e.g., summarizing a physician note).  Biomedical text processing exemplifies an emerging skill domain, but factual retrieval remains a challenge due to both factual gaps and errors (hallucinations) \cite{zhang2023siren}.

Precision medicine depends on the exact mapping of biomedical ontology terms to their machine-readable identifiers, ensuring that clinical and biomedical concepts are represented consistently across datasets \cite{rayan2021precision, robinson2015capturing}. This requirement exposes a potential weakness of current LLMs. Even advanced models such as GPT-4o achieve only 8\% accuracy when linking Human Phenotype Ontology (HPO) terms to their identifiers \cite{do2025mapping}.

Retrieval-augmented generation (RAG) and supervised fine-tuning have been proposed to address hallucinations and factual gaps~\cite{wang2023fine,hier2025simplified,shlyk2024real,thompson2023large}. Fine-tuning can inject knowledge~\cite{keloth2024advancing}, but it also carries risks, including knowledge degradation and the overwriting of previously learned facts~\cite{pletenev2025much}.  Even the best LLMs retrieve only about 65\% of the facts they know, and fine-tuning does not always help surface this latent knowledge~\cite{yuan2024towards}. Moreover, fine-tuning is most effective for facts that are partially known to the model rather than entirely novel~\cite{hier2025prior, speicher2024understanding}.  Fine-tuning performance follows scaling laws so that model size is more important than pretraining data size~\cite{zhang2024scaling}. 

Other research highlights additional limitations of fine-tuning, including over-reliance on rote memorization, limited generalization to new facts, and the \textit{reversal curse}, which reflects directional biases in knowledge acquisition~\cite{ovadia2024fine,berglund2023reversal,gekhman2024does,li2024gradual,yuan2024towards,wang2024generalization,wang2024re,wu2024finetunebench}. The balance between memorization and generalization varies systematically across task types: fact retrieval depends more on the memorization, whereas reasoning-intensive tasks rely on generalization during fine-tuning~\cite{wang2024generalization}.

A key distinction among biomedical terminologies is whether their identifiers are arbitrary codes or lexicalized symbols. As shown in Table~\ref{tab:identifier_examples}, terminologies such as HPO and the Gene Ontology (GO) use arbitrary identifiers that are by design opaque codes that are intentionally uninterpretable. In contrast, gene symbols are lexicalized, so that each symbol evokes the name of its referent term, allowing LLMs to leverage preexisting vector embeddings to associate terms with identifiers. When LLMs are fine-tuned on arbitrary codes, they are likely forced to rely on rote memorization of token sequences since these codes provide no intrinsic semantic cues. Lexicalized identifiers, by contrast, can enable a semantic alignment between terms and their machine codes, allowing LLMs to exploit preexisting embeddings rather than depending entirely on the memorization of token sequences. If this account is correct, lexicalized identifiers such as gene symbols~\cite{bruford2020guidelines} would be expected to support some generalization to unseen term–identifier pairs during fine-tuning, whereas arbitrary identifiers would not.

These theoretical distinctions motivated our design of our experiments. We created two fine-tuned models for each of three terminologies—HPO, GO, and gene symbol–protein mappings (referred to as GENE)—and fine-tuned each model bidirectionally (term$\rightarrow$identifier and identifier$\rightarrow$term). Prior work has shown that RAG-based approaches improve precision for low-prevalence terms~\cite{hier2025simplified,do2025ZS}, and that term popularity in the biomedical literature predicts the ability of LLMs to link terms to identifiers~\cite{do2025mapping,hier2025prior}. Building on these insights, we constructed frequency-balanced datasets for all three terminologies to isolate the relative influence of identifier popularity and identifier lexicalization on the success of supervised fine-tuning.
Two key patterns emerged: 
\begin{enumerate}
    \item The GO model efficiently memorized term–identifier pairs, whereas the HPO model struggled.
    \item The GENE model generalized to unseen term–identifier pairs, while the GO and HPO models failed to do so.
\end{enumerate}
We interpret these findings as reflecting two interacting factors: a \textit{popularity effect}, arising from the long-tail frequency distribution of biomedical terms, and a \textit{lexicalization effect}, reflecting the presence of semantically meaningful identifiers. Arbitrary identifiers such as those found in GO and HPO constrain models to rote memorization, and rare identifiers (as in HPO) further limit memorization efficiency. In contrast, lexicalized identifiers, such as gene symbols, enable semantic alignment that supports limited generalization to unseen mappings. This framework provides systematic predictions of when fine-tuning will succeed or fail in linking biomedical terms to their identifiers.

\begin{table*}[t!]
\centering
\caption{Examples of arbitrary and {lexicalized} biomedical identifiers.}
\label{tab:identifier_examples}
\small
\begin{tabular}{llll}
\toprule
\textbf{Terminology} & \textbf{Term} & \textbf{Identifier} & \textbf{Identifier Type} \\
\midrule
GO   & nucleus                  & GO:0005634 & arbitrary \\
GO   & cytosol                  & GO:0005829 & arbitrary \\
HPO  & tremor                   & HP:0001337 & arbitrary \\
HPO  & ataxia                   & HP:0001251 & arbitrary \\
GENE & fibroblast growth factor & FGF        & lexicalized \\
GENE & breast cancer 1          & BRCA1      & lexicalized \\
GENE & tumor protein p53        & TP53       & lexicalized \\
GENE & superoxide dismutase 1   & SOD1       & lexicalized \\
\bottomrule
\end{tabular}
\begin{tablenotes}
\item  \footnotesize GO = Gene Ontology (cellular component hierarchy);  HPO = Human Phenotype Ontology 
\item  GENE = shorthand for protein names linked to gene symbols
\item  {Arbitrary identifiers do not resemble the linked term. Lexicalized identifiers are designed to evoke the linked term.}
\end{tablenotes}
\end{table*}

\begin{figure}
    \centering    \includegraphics[width=0.70\linewidth]{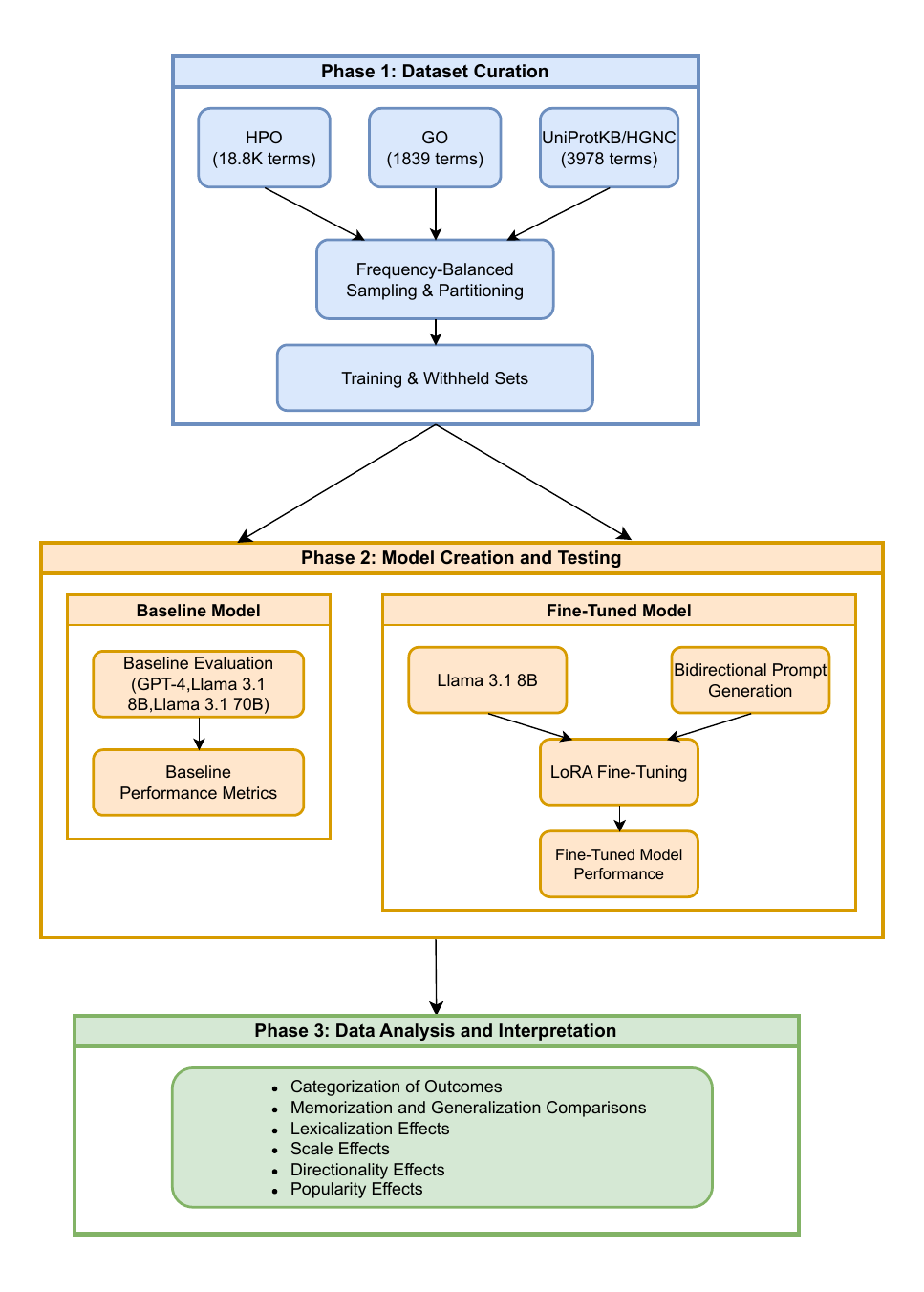}
    \caption{Overview of Methods. Phase 1: frequency-balanced train and validation test sets were created from three biomedical terminologies. Phase 2: Fine-tuned models were created for each terminology and performance compared to baseline models. Phase 3: Gains from fine-tuning were attributed to either memorization or generalization. Effects of scale, popularity, directionality, and lexicalization were assessed.}
    \label{fig:project-workflow}
\end{figure}

\section*{Methods}
We evaluated the ability of large language models to link biomedical terms to their identifiers across three widely used terminologies: HPO \cite{hpo_2012}, GO \cite{go_2023}, and UniProtKB protein names \cite{uniprot_2015} with their corresponding HGNC gene symbols \cite{hgnc_2006} (\textbf{GENE}). We investigated term–identifier linking in both directions: term $\rightarrow$ identifier and identifier $\rightarrow$ term. Figure~\ref{fig:project-workflow} summarizes the research workflow.

\subsection*{Datasets}
HPO contains 18,800 terms describing the phenotypic abnormalities of human diseases. Its primary use is annotating the signs and symptoms of rare disorders, with more than 200,000 annotations that link diseases to their signs~\cite{hpo_annotations}. HPO uses arbitrary identifiers.

GO has 43,303 concepts organized into three hierarchies: biological process, molecular function, and cellular component. In this study, we focused on 1,839 terms from the cellular component (CC) hierarchy. GO (CC) provides a standardized terminology for describing the locations of proteins in the cell and adheres to OBO Foundry principles~\cite{obo_foundry_2021}. Across species, GO includes over 9 million annotations~\cite{go_annotations}. Like HPO, GO employs arbitrary identifiers.

UniProtKB protein names and HGNC gene symbols are interlocking terminologies that link gene symbols, gene names, and protein names. For simplicity, we refer to these interlocking terminologies as \textbf{GENE}. The UniProtKB/Swiss-Prot terminology includes approximately 20,000 human proteins, each associated with an HGNC gene symbol~\cite{uniprot_human_proteome, uniprot_data}. Gene symbols are intentionally chosen to align with their corresponding gene and protein names \cite{wain2002guidelines}.

The three terminologies differ in structure, size, adoption, and frequency of use in the biomedical literature. HPO and GO are hierarchical ontologies and are distributed in OWL, JSON, and OBO formats via the OBO Foundry~\cite{obo_foundry_website} and in CSV format via NCBO BioPortal~\cite{bioportal}. HPO and GO are organized hierarchically, whereas the UniProtKB/HGNC (GENE) terminologies are flat. Together, they provide a robust testbed for evaluating the memorization and generalization of term–identifier pairs during fine-tuning. Importantly, their contrasting identifier characteristics, arbitrary codes (HPO and GO) versus lexicalized symbols (GENE), provide a direct test of our hypothesis that lexicalization supports generalization. To ensure a comparable evaluation across terminologies of different sizes and frequencies, we implemented a stratified sampling strategy.

\subsection*{Sampling Strategy}
Most biomedical terminologies exhibit a pronounced \textit{long-tail} distribution consistent with Zipf’s Law~\cite{zipf2012human}: a small number of head terms occur frequently, while the majority are rare. LLMs often struggle with these long-tail terms, which receive little exposure during pretraining~\cite{shen2015long,zhang2025systematic,yi2025can,hier2025prior}. Because it is infeasible to fine-tune on every term, we constructed frequency-balanced datasets for each terminology (HPO, GO, GENE) that included terms sampled from the head, body, and tail of the frequency distribution~\cite{do2025mapping,do2024retrievalthesis}. 
Identifier usage counts were retrieved from PubMed Central (PMC) via the PMC API \cite{beck2003pubmed}. For each terminology, identifiers were ranked by frequency, partitioned into twenty equal-sized bins, and ten term–identifier pairs were randomly sampled from each bin. Because terms often appear in the biomedical literature without their identifiers, identifier frequency provides a more reliable indicator of long-tail behavior than term frequency alone~\cite{do2024retrievalthesis,hier2025prior}. 
This stratified sampling ensured balanced representation across the frequency spectrum and produced 200 pairs for each terminology (HPO, GO, and GENE). These curated datasets formed the foundation for baseline model evaluation before fine-tuning.

\subsection*{Baseline Evaluation}
We evaluated baseline performance on unmodified Llama 3.1 models (8B and 70B parameters)~\cite{llama_together_ai} and GPT-4o~\cite{gpt4_openai}. Each model was tested on the same term–identifier sets using identical prompt templates. This evaluation reflects the intrinsic biomedical knowledge encoded during pretraining and establishes a baseline for subsequent fine-tuning. 
Performance was assessed in both mapping directions (term~$\rightarrow$~identifier and identifier~$\rightarrow$~term) across all three datasets (HPO, GO, and GENE). Mapping accuracy was measured using \texttt{top-1 accuracy} (hits@1), where a prediction was scored as correct if the model’s top-ranked output exactly matched the ground truth identifier (or term).  With these baselines established, we next applied parameter-efficient fine-tuning to test whether domain-specific knowledge could be acquired without full model retraining.

\begin{table}[t!]
\captionsetup{skip=3pt}
\caption{Prompt templates for term–identifier mapping fine-tuning.}
\label{tab:prompts}
\centering
\begin{tabular}{@{}ll@{}}
\toprule
\textbf{Prompt} & \textbf{Template} \\
\midrule
1 & What is the [ONTOLOGY] identifier for the [ONTOLOGY] term [TERM]? \\
2 & The [ONTOLOGY] term [TERM] has what [ONTOLOGY] identifier? \\
  & Respond only with the [ONTOLOGY] identifier. \\
3 & Provide the [ONTOLOGY] identifier for: [TERM] \\
4 & What is the ontology identifier for [TERM] in [ONTOLOGY]? \\
5 & Return only the [ONTOLOGY] identifier for the term: [TERM] \\
\bottomrule
\end{tabular}
\begin{tablenotes}[para,flushleft] \footnotesize
Each term–identifier pair was expanded into five distinct prompts to capture variation in phrasing. [ONTOLOGY] is the placeholder for the terminology, and [TERM] is the placeholder for the term being probed. Reverse mapping prompts mirrored the forward prompts, substituting identifiers for terms.
\end{tablenotes}
\end{table}

\subsection*{Parameter-Efficient Fine-Tuning}
For each terminology (HPO, GO, and GENE), equal-sized training sets (200 term–identifier pairs) and validation sets (unseen terms) were constructed. Each training pair was expanded into five distinct prompts to expose the LLM to varied phrasing, as shown in Table~\ref{tab:prompts}. Reverse mapping prompts mirrored the forward prompts, substituting identifiers for terms. This produced 1,000 training prompts per terminology in each direction.

We performed fine-tuning on the Llama 3.1 8B Instruct model \cite{llama31_together}. Parameter-efficient adaptation used Low-Rank Adaptation (LoRA; rank $r=64$, scaling $\alpha=128$) \cite{lora_2021}, applied to all linear transformer layers. Optimization employed supervised fine-tuning with a learning rate of $1\times10^{-5}$ (cosine scheduler, no warm-up, cycle length = 0.5), batch size = 32, max gradient norm = 1, and no weight decay for a total of 20 epochs. The \texttt{train-on-inputs} option was set to \texttt{auto}. Separate models were trained for each terminology and mapping direction, yielding six fine-tuned models. Validation loss plateaued after $\sim$5 epochs for HPO and GENE mappings, and $\sim$15 epochs for GO. We categorized outputs from the fine-tuned models using the following evaluation framework.

\subsection*{Evaluation Framework}

Fine-tuning performance was evaluated by comparing accuracy before and after fine-tuning for each test instance (term–identifier or identifier–term pair). Each instance was assigned to one of four mutually exclusive categories: (i) \textit{Gainer}: terms that were \textit{incorrect} at baseline but \textit{correct} after fine-tuning and indicating that knowledge was acquired. (ii) \textit{Loser}: terms that were \textit{correct} at baseline but \textit{incorrect} after fine-tuning indicating that knowledge was degraded. (iii) \textit{Correct}: terms that were \textit{correct} at baseline and after fine-tuning indicating that knowledge was preserved. (iv) \textit{Incorrect}: terms that were \textit{incorrect} at baseline and after fine-tuning. This framework distinguishes gains, losses, and stable outcomes, providing a comprehensive view of fine-tuning effects across terminologies. 

We evaluated the performance of the model based on three key dimensions: memorization vs. generalization, lexicalization, and popularity.

\subsection*{Memorization vs. Generalization}

Definitions of memorization and generalization in LLM fine-tuning vary across studies~\cite{ni2025training,wang2024generalization,wang2024re}. Following Bridgman’s notion of \textit{operational definitions}~\cite{bridgman1927logic}, we define these terms strictly by their empirical manifestations. 

\textit{Memorization}: Accuracy gains on mappings from the \textit{training set} (term–identifier pairs explicitly seen during fine-tuning). 

\textit{Generalization}: Accuracy gains on mappings from the \textit{validation set} (term–identifier pairs likely encountered during pretraining but not seen during fine-tuning).

These operational definitions are broadly consistent with recent formal definitions of generalization~\cite{wu2025rote}. 

\subsection*{Lexicalization}

To probe why some terminologies support generalization while others do not, we also examine the role of lexicalization. \textit{Lexicalization} refers here to the degree to which an identifier encodes semantic information that evokes its associated term. Lexicalized identifiers share lexical overlap with the term itself (e.g., \texttt{TP53} $\leftrightarrow$ \texttt{tumor protein p53}), whereas an arbitrary identifier such as \texttt{HP:0001250} $\rightarrow$ \texttt{seizure} does not. In this study, we define lexicalization as the presence of semantic alignment between the embeddings of a term and its corresponding identifier. To test for lexicalization, we used Llama 3.1 8B to generate embeddings and then compared the embeddings of terms with those of their identifiers.

We used the 4,096-dimensional embeddings from the Llama 3.1 8B model (\texttt{meta-llama/Meta-Llama-3-8B}). Each string (term or identifier) was tokenized, final-layer hidden states were extracted, and token vectors were averaged to yield a single vector. Cosine similarity was computed for each true term–identifier pair and compared with similarities to non-matching identifiers. Higher similarity for matched pairs was interpreted as evidence of lexicalization.

As a complementary measure, all embedding vectors ($n=600$) were projected into two dimensions using principal component analysis (PCA). Euclidean distances between terms and their paired identifiers were compared with distances to non-paired identifiers. Reduced distances between true pairs were interpreted as additional evidence of lexicalization. This visualization tested whether LLM embeddings organize terms and identifiers within a shared semantic space. Finally, to examine how prior exposure might shape these outcomes, we analyzed proxies for popularity with each term–identifier pair.

\subsection*{Popularity}

The frequency with which a fact appears in the training corpus has been termed \textit{popularity} and has been shown to predict both its retrievability and its ease of consolidation during fine-tuning~\cite{yuan2024towards, chang2024large}. Facts with higher popularity are easier to recall, whereas rare, long-tail facts are less accessible to the model. In earlier work, we referred to this property as \textit{familiarity} and hypothesized that model familiarity with specific terms and identifiers during pretraining would facilitate fine-tuning success~\cite{hier2025prior}. Here, we adopt the community-standard term \textit{popularity} to emphasize its correspondence with frequency-based effects observed across multiple studies. Because the exact frequencies of facts in pretraining corpora are unknown, we approximate popularity using the following proxies:

\begin{enumerate}
\item \textit{Identifier frequency in biomedical text:} counts of HPO identifiers, GO identifiers, and gene symbols in the PubMed Central (PMC) full-text database~\cite{beck2003pubmed}.
\item \textit{Term frequency in biomedical text:} counts of HPO terms, GO terms, and protein names in PMC.
\item \textit{Annotation frequency in curated resources:} counts of HPO terms used in disease annotation~\cite{hpo_annotations}, counts of GO cellular component terms used for annotations~\cite{go_annotations}, and counts of protein annotations in the UniProtKB dataset~\cite{uniprot_data}.
\end{enumerate}

These three measures serve as operational proxies for popularity, reflecting the likelihood that a model encountered a given term–identifier association during pretraining. Popularity measures (annotation counts, identifier counts in PMC, and term counts in PMC) were Laplace-smoothed by adding one to each value and then $\log_{10}$-transformed to reduce skewness and stabilize variance. The transformed variables were analyzed using two-way analyses of variance (ANOVA) with \textit{Terminology} (HPO, GO, GENE) and \textit{Correctness} (Match = 0 or 1) as fixed factors. When the ANOVA revealed significant main effects, pairwise differences among ontologies were examined using the Games–Howell post hoc test, which does not assume equal variances or sample sizes.

\section*{Results}\label{sec2}

We evaluated baseline performance across three models (Llama 3.1 8B, Llama 3.1 70B, and GPT-4o) on all three terminologies and in both mapping directions. The fine-tuning experiments used the Llama 3.1 8B model exclusively.

\begin{table}[t!]
\centering
\caption{Baseline and fine-tuned performance across the biomedical terminologies and mapping directions.}
\label{tab:results_by_model_scale}
\begin{tabular}{lrrrrrr}
\toprule
\textbf{Mapping} & \textbf{Llama 8B} & \textbf{Llama 8B FT} & \textbf{$\Delta$ FT} & \textbf{Llama 70B} & \textbf{GPT-4o} \\
\midrule
HPO identifier $\rightarrow$ Term          & 0.0 & 0.0 & 0.0  & 0.0  & 0.2 \\
HPO Term $\rightarrow$ identifier          & 0.4 & 0.6 & +0.2 & 4.7  & 8.8 \\
GO identifier $\rightarrow$ Term           & 0.5 & 4.0 & +3.5 & 7.2  & 18.4 \\
GO Term $\rightarrow$ identifier           & 2.4 & 9.8 & +7.4 & 17.5 & 35.6 \\
Gene $\rightarrow$ Protein         & 22.8 & 32.6 & +9.8 & 51.6 & 59.7 \\
Protein $\rightarrow$ Gene         & 73.7 & 77.0 & +3.3 & 88.9 & 95.3 \\
\bottomrule
\end{tabular}
\begin{tablenotes}[para,flushleft] \footnotesize
 Accuracy values are percentages. $\Delta$ FT is the calculated performance gain from fine-tuning Llama 3.1 8B. All models were evaluated on both training and validation terms. 
\end{tablenotes}
\end{table}

\begin{figure}
    \centering
    \includegraphics[width= 0.98 \linewidth]{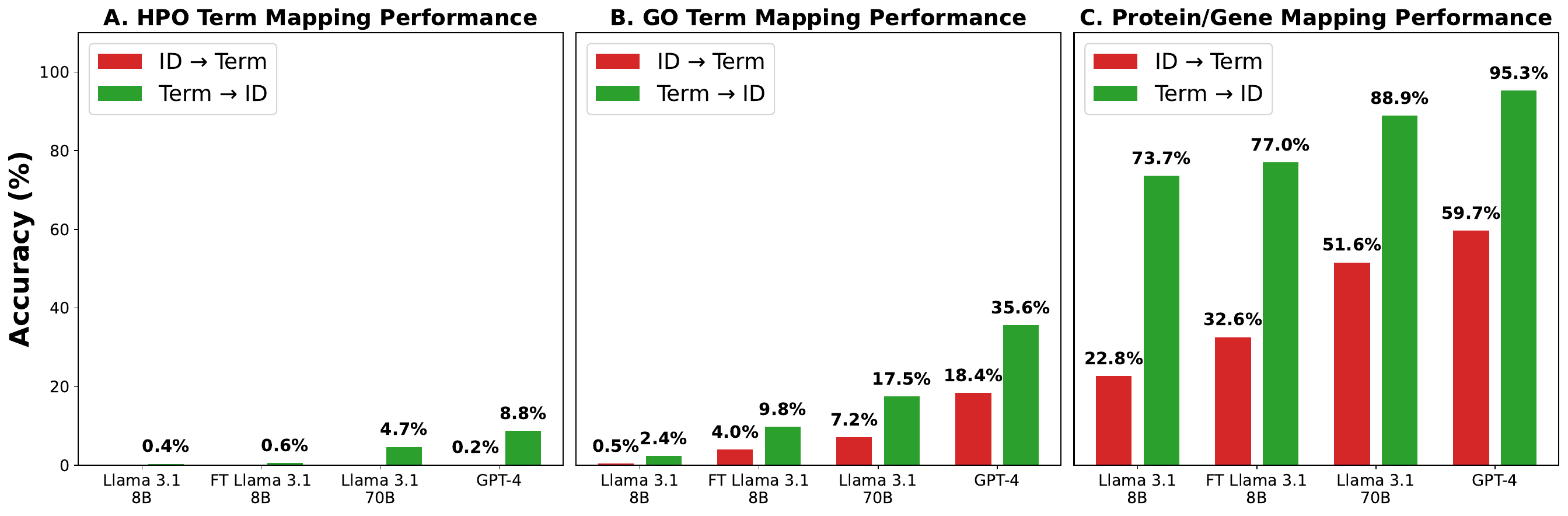}
    \caption{Bidirectional term mapping performance comparison across biomedical terminologies and model scales. Overall, Term$\rightarrow$identifier mappings perform better than identifier$\rightarrow$term, with Protein/Gene achieving the highest accuracy, GO moderate, and HPO the poorest across model scales. Performance evaluated on combined training and validation sets.}
    \label{fig:model_comparison}
\end{figure}

\subsection*{Baseline Performance}
Table~\ref{tab:results_by_model_scale} presents baseline accuracy for all models and terminologies. GPT-4o achieved the highest accuracy, followed by Llama 3.1 70B and Llama 3.1 8B. The GENE model showed the highest accuracy (22.8--95.3\%), the GO model showed intermediate accuracy (0.5--35.6\%), and the HPO model showed the lowest accuracy (0.0--8.8\%). 

\subsection*{Effect of Fine-Tuning on Accuracy}
Fine-tuning effects for Llama 3.1 8B varied by terminology (Table~\ref{tab:results_by_model_scale}). Accuracy gains ($\Delta$FT) were substantial for the GO and GENE models but minimal for the HPO model. Even after fine-tuning, performance remained below that of the baseline Llama 3.1 70B and GPT-4o models.

\subsection*{Directionality Effects}
Figure~\ref{fig:model_comparison} depicts accuracy by mapping direction. Across all models and terminologies, term~$\rightarrow$~identifier accuracy exceeded identifier~$\rightarrow$~term accuracy, a pattern that held in both baseline and fine-tuned conditions. This directional asymmetry was pronounced across all three systems: GPT-4o achieved 95.3\% for protein~$\rightarrow$~gene versus 59.7\% for the reverse direction, 35.6\% for GO term~$\rightarrow$~identifier versus 18.4\% for identifier~$\rightarrow$~term, and 8.8\% for HPO term~$\rightarrow$~identifier versus 0.2\% for identifier~$\rightarrow$~term.

\begin{table*}[t!]
\centering
\caption{Assessment of fine-tuning effectiveness on biomedical bidirectional mapping performance.}
\label{tab:all_terminology_performance}
\small
\begin{tabular}{lllrr}
\toprule
\textbf{Terminology} & \textbf{Direction} & \textbf{Category} & \textbf{Validation \%} & \textbf{Trained \%} \\
&&&(Unseen)&(Seen)\\
\midrule
\multirow{8}{*}{HPO} & \multirow{4}{*}{identifier $\rightarrow$ Term} & Gainer & 0.0 & 0.0 \\
& & Loser & 0.0 & 0.0 \\
& & Correct & 0.0 & 0.0 \\
& & Incorrect & 100.0 & 100.0 \\
\cmidrule{2-5}
& \multirow{4}{*}{Term $\rightarrow$ identifier} & Gainer & 0.3 & 2.5 \\
& & Loser & 0.2 & 0.0 \\
& & Correct & 0.3 & 0.5 \\
& & Incorrect & 99.3 & 97.0 \\
\midrule
\multirow{8}{*}{GO} & \multirow{4}{*}{identifier $\rightarrow$ Term} & Gainer & 0.2 & 33.0 \\
& & Loser & 0.4 & 0.0 \\
& & Correct & 0.2 & 0.5 \\
& & Incorrect & 99.2 & 66.5 \\
\cmidrule{2-5}
& \multirow{4}{*}{Term $\rightarrow$ identifier} & Gainer & 0.7 & 77.0 \\
& & Loser & 1.8 & 0.0 \\
& & Correct & 0.6 & 2.5 \\
& & Incorrect & 96.8 & 20.5 \\
\midrule
\multirow{8}{*}{GENE} & \multirow{4}{*}{Gene $\rightarrow$ Protein} & Gainer & 13.9 & 48.0 \\
& & Loser & 6.0 & 3.5 \\
& & Correct & 16.7 & 22.5 \\
& & Incorrect & 63.4 & 26.0 \\
\cmidrule{2-5}
& \multirow{4}{*}{Protein $\rightarrow$ Gene} & Gainer & 7.0 & 24.0 \\
& & Loser & 4.6 & 1.5 \\
& & Correct & 69.2 & 69.5 \\
& & Incorrect & 19.2 & 5.0 \\
\bottomrule
\end{tabular}
\begin{tablenotes}[para,flushleft] \footnotesize
Assessment of fine-tuning effectiveness on biomedical bidirectional mapping performance (Training: 200 pairs per terminology; Validation: HPO $\sim$18,600, GO $\sim$1,639, GENE $\sim$3,778 terms). Results are separated into four outcome categories (defined in Methods) and reported for both Trained (Seen) and Validation (Unseen) terms. This breakdown allows comparison of memorization (Seen) and generalization (Unseen) across terminologies.
\end{tablenotes}
\end{table*}

\begin{figure}
    \centering
    \includegraphics[width= 0.98 \linewidth]{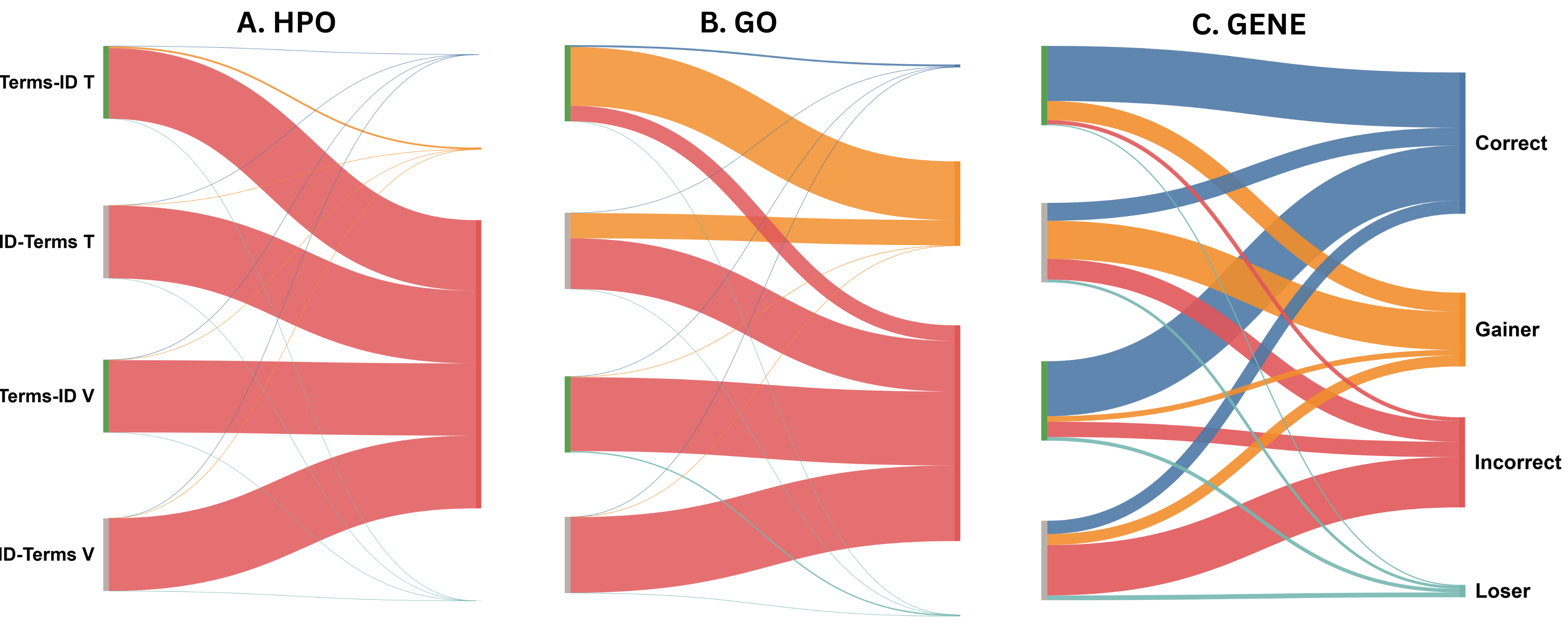}
    \caption{Sankey diagrams illustrating fine-tuning outcomes across HPO, GO, and GENE models.
Color flows visualize how terms transition after fine-tuning: blue = correct, orange = gainers (newly correct), red = incorrect, and green = degraded.
Panel A (HPO) shows dominance of red flows, reflecting the long tail of low-popularity identifiers that remain unmapped after fine-tuning.
Panel B (GO) exhibits strong orange flows, consistent with moderate popularity and effective memorization of trained terms.
Panel C (GENE) displays extensive blue and orange flows, indicating high-popularity, lexicalized identifiers that support both memorization and generalization to unseen terms.
Green (degradation) flows are visible mainly in the GENE model, marking the upper limit of representational saturation.}
    \label{fig:all_sankey}
\end{figure}

\subsection*{Knowledge Gains and Losses Following Fine-Tuning}

Table~\ref{tab:all_terminology_performance} classifies each mapping into four mutually exclusive outcomes after fine-tuning: \textit{Gainer}, \textit{Loser}, \textit{Correct}, and \textit{Incorrect}. Results are separated by trained terms (seen during fine-tuning) and validation terms (unseen). These outcomes are also visualized as Sankey flows (Figure~\ref{fig:all_sankey}).

For the HPO model, almost all mappings (97.0--100\%) remained \textit{Incorrect}. Gainers comprised only 0.0--2.5\% of trained mappings and 0.0--0.3\% of validation mappings.

For the GO model, strong memorization occurred on trained terms, with Gainers comprising 33.0--77.0\% of trained mappings. In contrast, validation mappings showed minimal gains (0.2--0.7\%). GO term~$\rightarrow$~identifier mappings showed 77.0\% Gainers in the trained set versus 0.7\% in the validation set.

For the GENE model, substantial gains occurred for both trained and validation terms. Gainers comprised 24.0--48.0\% of trained mappings and 7.0--13.9\% of validation mappings, while Losers comprised 1.5--3.5\% of trained mappings and 4.6--6.0\% of validation mappings.

\subsection*{Derived Metrics of Fine-Tuning Performance}

Table~\ref{tab:derived_metrics} summarizes memorization (gains on trained terms), generalization (gains on unseen terms from the validation set), degradation (losses after fine-tuning), and overall accuracy across terminologies.

The HPO model showed virtually no improvement, with minimal memorization (0.0–2.5\%) and negligible generalization (0.0–0.3\%). The GO model demonstrated strong memorization on trained (seen) terms, reaching up to 77\% for term~$\rightarrow$~identifier mappings, but showed minimal generalization ($\leq$1\%). In contrast, the GENE model exhibited both robust memorization (24–48\%) and notable generalization on the validation set (7–14\%), although it was accompanied by some degradation (4–10\%). Fine-tuning consistently improved performance on trained terms across terminologies, but only the GENE model showed meaningful success on unseen terms.

\begin{table*}[t!]
\centering
\caption{Derived performance metrics for fine-tuned Llama 3.1 8B across the three biomedical terminologies.}
\label{tab:derived_metrics}
\small
\begin{tabular}{lrrrr}
\toprule
\textbf{Task} & \textbf{Memorized} & \textbf{Generalized} & \textbf{Degraded} & \textbf{Accuracy*} \\
\midrule
HPO identifier $\rightarrow$ Term & 0.0 & 0.0 & 0.0 & 0.0 \\
HPO Term $\rightarrow$ identifier & 2.5 & 0.3 & 0.2 & 3.0 \\
GO identifier $\rightarrow$ Term & 33.0 & 0.2 & 0.4 & 33.5 \\
GO Term $\rightarrow$ identifier & 77.0 & 0.7 & 1.8 & 79.5 \\
Gene $\rightarrow$ Protein & 48.0 & 13.9 & 9.5 & 70.5 \\
Protein $\rightarrow$ Gene & 24.0 & 7.0 & 6.1 & 87.5 \\
\bottomrule
\end{tabular}
\begin{tablenotes} \footnotesize
\item All values are percentages.  * Accuracy is for trained terms only. Accuracy = Correct + Gainer – Loser 
 \item Memorized = gains on trained terms; \; \; Generalized = gains on validation terms; 
 \item Degraded = losses on trained or validation terms.  Values derived from Table~\ref{tab:all_terminology_performance}
\end{tablenotes}
\end{table*}

\begin{figure}
    \centering
    \includegraphics[width= 0.98 \linewidth]{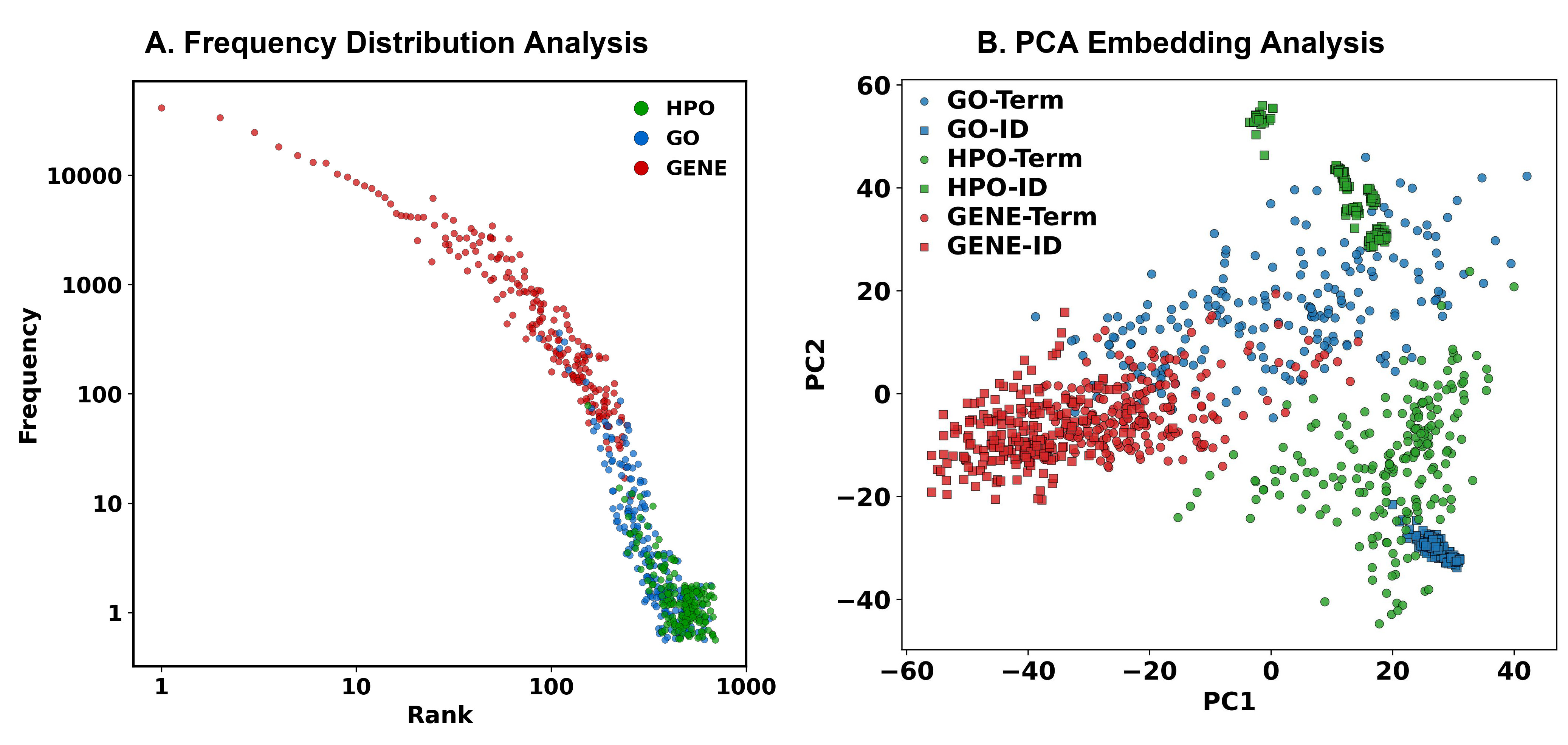}
    \caption{Identifier frequency and semantic alignment patterns across biomedical terminologies based on the 200 trained terms only. \textbf{(A)} Log-log rank-frequency plot of identifiers in PubMed Central showing GENE symbols (red circles) in the distribution head, GO identifiers (blue circles) in the body, and HPO identifiers (green circles) in the long tail. \textbf{(B)} PCA projection of Llama 3.1 8B embeddings showing overlapping clusters for GENE terms (red circles) and identifiers (red squares) versus clear separation between terms (circles) and identifiers (squares) for GO (blue)  and HPO (green).}
    \label{fig:combined_freq_embed}
\end{figure}

\begin{figure}
    \centering
    \includegraphics[width= 0.98 \linewidth]{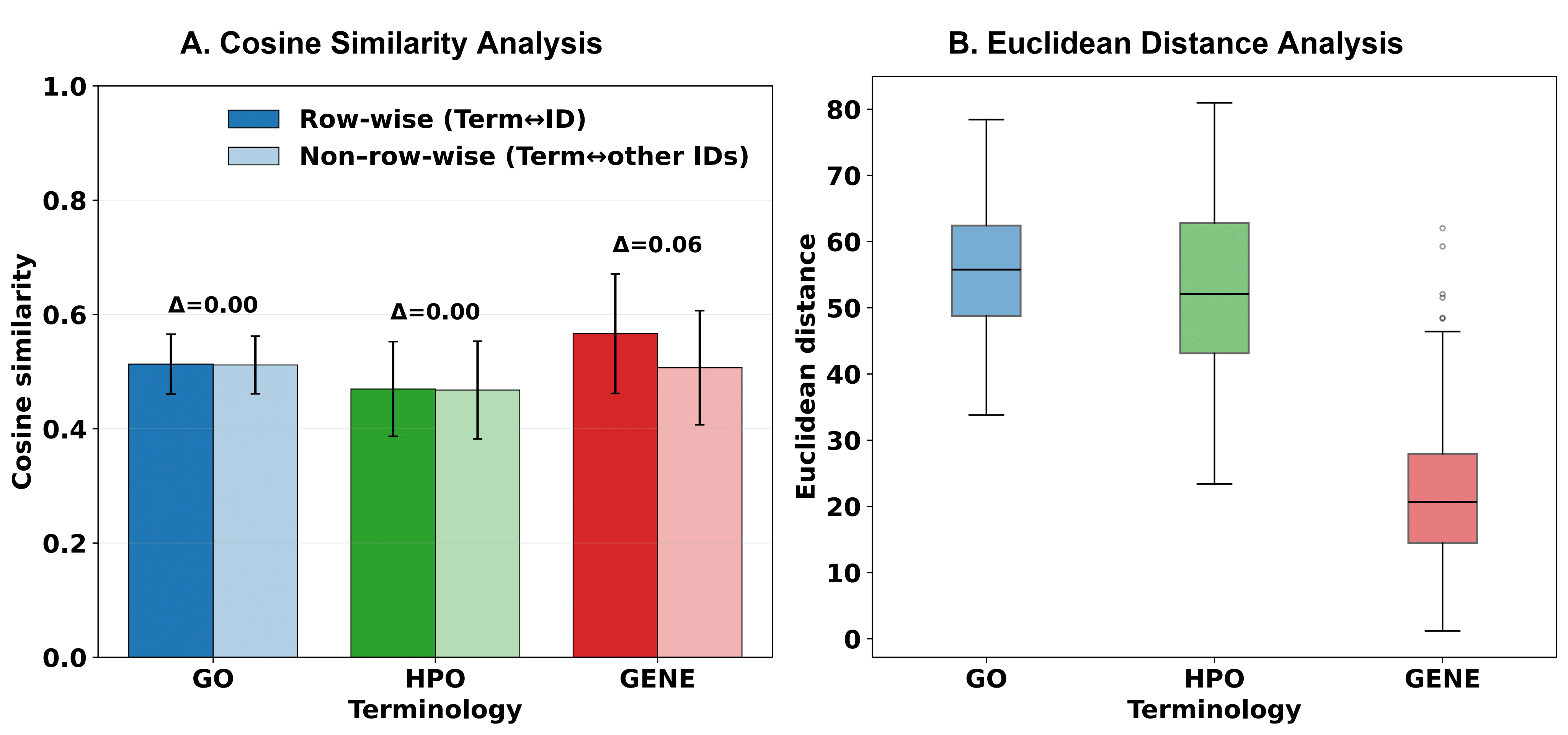}
    \caption{Semantic alignment between terms and identifiers across biomedical terminologies based on the trained terms only. \textbf{(A)} Row-wise vs.~non–row-wise cosine similarity reveals semantic alignment only for GENE term--identifier pairs (red bars). Bars show mean cosine similarity ($\pm$SD) between each term and its own identifier (solid color) versus the same term paired with all other identifiers in the same terminology (lighter shade). \textbf{(B)} Euclidean distances between terms and identifier in PCA space (4096 $\rightarrow$ 2 dimensions). GENE term–identifier pairs (red box) are close to each other, unlike GO and HPO.}
    \label{fig:semantic_alignment}
\end{figure}

\subsection*{Lexicalization}

PCA projection of 4,096-dimensional embeddings (Figure~\ref{fig:combined_freq_embed}B) showed distinct spatial patterns across terminologies. GENE terms and identifiers formed overlapping clusters. HPO and GO identifiers formed separate clusters distinct from their corresponding terms, suggesting that the model does not align textual descriptors with machine-readable identifiers for these ontologies. This analysis tests whether Llama embeddings capture term–identifier correspondence in their internal representational space.

Cosine similarity analysis (Figure~\ref{fig:semantic_alignment}A) showed higher row-wise similarity (term paired with its own identifier) versus non-row-wise similarity (term paired with other identifiers) for GENE only. GENE row-wise mean: 0.54 (SD = 0.10); non-row-wise mean: 0.48 (SD = 0.08); difference: $\Delta$ mean = +0.060; Welch’s t = 8.05, p $<$ 10$^{-6}$, n = 200. Although the absolute difference was modest (0.06), it represents a 12.5\% relative increase over the baseline mean (0.48), underscoring that the alignment effect is statistically robust and practically meaningful. HPO and GO showed $\Delta$ mean = 0.00, indicating no alignment between terms and their identifiers.

Euclidean distance in PCA space (Figure~\ref{fig:semantic_alignment}~B) confirmed this pattern, showing that GENE term–identifier pairs exhibited smaller distances than HPO or GO pairs.

\begin{table}[t!]
\centering
\caption{Popularity Proxies by Terminology}
\label{tab:fact_popularity}
\begin{tabular}{ll
                c
                S[table-format=5.1]
                S[table-format=5.1]}
\toprule
\textbf{Measure} & \textbf{Group} & \textbf{Category (N)} &
\multicolumn{1}{r}{\textbf{Mean}} &
\multicolumn{1}{r}{\textbf{SD}} \\
\midrule
\multirow{6}{*}{PMC identifier} 
& \multirow{2}{*}{HPO}  & Correct (0)     & 0.0    & 0.0 \\
&                       & Incorrect (200) & 1.1    & 5.7  \\
\cmidrule(lr){2-5}
& \multirow{2}{*}{GO}   & Correct (53)    & 24.9   & 84.2 \\
&                       & Incorrect (147) & 14.7   & 40.5 \\
\cmidrule(lr){2-5}
& \multirow{2}{*}{GENE} & Correct (141)   & 2063.2 & 5361.0 \\
&                       & Incorrect (59)  & 1362.5 & 2717.2 \\
\midrule
\multirow{6}{*}{Annotations} 
& \multirow{2}{*}{HPO}  & Correct (0)     & 0.0    & 0.0 \\
&                       & Incorrect (200) & 20.4   & 133.3 \\
\cmidrule(lr){2-5}
& \multirow{2}{*}{GO}   & Correct (53)    & 715.8  & 4852.7 \\
&                       & Incorrect (147) & 68.3   & 252.0 \\
\cmidrule(lr){2-5}
& \multirow{2}{*}{GENE} & Correct (141)   & 37.1   & 20.7 \\
&                       & Incorrect (59)  & 31.6   & 13.9 \\
\midrule
\multirow{6}{*}{PMC term counts} 
& \multirow{2}{*}{HPO}  & Correct (0)     & 0.0     & 0.0 \\
&                       & Incorrect (200) & 23451.7 & 93838.4 \\
\cmidrule(lr){2-5}
& \multirow{2}{*}{GO}   & Correct (53)    & 7033.9  & 35379.5 \\
&                       & Incorrect (147) & 7602.9  & 45311.3 \\
\cmidrule(lr){2-5}
& \multirow{2}{*}{GENE} & Correct (141)   & 2184.1  & 6109.4 \\
&                       & Incorrect (59)  & 1094.9  & 4058.7 \\
\bottomrule
\end{tabular}
\begin{tablenotes}[para,flushleft] \footnotesize
Counts are shown in parentheses (N).
Means are $\pm$ standard deviations (SD).
\end{tablenotes}
\end{table}

\sisetup{
  table-align-text-post=false,
  table-number-alignment=center,
  table-figures-integer=5,
  table-figures-decimal=1
}

\subsection*{Popularity}

Descriptive statistics for the three proxies of popularity are presented in Table~\ref{tab:fact_popularity}. Mean values are shown for PubMed Central (PMC) identifier counts, annotation counts, and PMC term counts, grouped by whether the model correctly or incorrectly performed the term~$\rightarrow$~identifier mapping task. 
For both PMC identifier counts and annotation counts, correctly mapped terms exhibited higher values—indicating greater popularity—than incorrectly mapped terms. This relationship was less consistent for PMC term counts, suggesting that raw term frequency alone does not fully capture model familiarity.

Among the three ontologies, PMC identifier counts differed sharply, following a clear ordering of \textbf{GENE $>$ GO $>$ HPO}. The mean identifier frequency in PMC was approximately 200-fold higher for GENE identifiers (gene symbols) than for HPO identifiers, with GO occupying an intermediate position. This pattern was confirmed by two-way ANOVA and Games–Howell post hoc tests, which showed significant pairwise differences among all three ontologies (all $p < 0.001$). The strong gradient in PMC identifier counts indicates that models were likely exposed far more frequently to gene symbols than to GO or HPO identifiers during pretraining—consistent with the hypothesis that greater corpus exposure enhances representational strength and retrieval accuracy.

\section*{Discussion}
This study evaluated the ability of large language models to learn biomedical term–identifier mappings across three representative terminologies (HPO, GO, and GENE) and three LLM models (Llama 3.1 models (8B and 70B parameters) and GPT-4o~.
Several consistent patterns emerged from the results. First, model scale effects prior to fine-tuning were evident: larger models consistently outperformed smaller ones. 
Second, directionality effects were robust: mapping terms to identifiers was more accurate than mapping identifiers to terms across all models and terminologies. 
Third, the effectiveness of fine-tuning varied across terminologies: fine-tuning substantially improved GO and GENE mappings, but yielded negligible gains for HPO.
Finally, only the GENE terminology demonstrated evidence of generalization, in which the model surfaced unseen term–identifier pairs after fine-tuning. By contrast, improvements for HPO and GO were restricted to memorization of training examples. We interpret these findings in light of four explanatory factors: model scale, directionality, popularity, and lexicalization.

\subsection*{Model Scale Effects}
Across all tasks, larger models outperformed smaller ones, with GPT-4o achieving the highest baseline accuracy, followed by Llama 3.1 70B and Llama 3.1 8B (Table ~\ref{tab:results_by_model_scale}). Fine-tuning improved the 8B model relative to its own baseline, but it never surpassed the performance of the larger pretrained models. This pattern reflects the well-documented scaling laws of LLMs~\cite{kaplan2020scaling, lu2024scaling, zhang2024scaling}, where greater parameter counts, larger embedding dimensions, and more extensive pretraining exposure yield stronger performance. Even with parameter-efficient fine-tuning, architectural capacity and pretraining scale remain the dominant determinants of success. 

\subsection*{Directionality Effects}

Another robust pattern was the consistent directionality effect: mapping terms to identifiers 
was always more accurate than mapping identifiers to terms, regardless of model size or 
terminology (Figure~\ref{fig:model_comparison}). We hypothesize that this asymmetry is an inherent artifact of the 
autoregressive training objective of LLMs. During pretraining, models are optimized to 
predict the next token given a preceding context. In biomedical corpora, natural language 
terms usually precede their identifiers (e.g., \textit{ataxia} $\rightarrow$ HP:0001251). 
Thus, the forward direction (term~$\rightarrow$~identifier) aligns with the statistical 
structure of pretraining data, while the reverse direction does not. 

This explanation is consistent with prior observations of order sensitivity in LLM reasoning tasks, including the so-called “reversal curse” \cite{berglund2023reversal}, where models trained on ``A is B'' fail to infer the logically equivalent ``B is A.'' Applied here, the phenomenon suggests that identifiers are not encoded as flexible semantic units, but rather as fixed token sequences whose predictability depends strongly on their position relative to terms.  We next examine how popularity and frequency distributions across terminologies 
influence the success or failure of fine-tuning.

\subsection*{Popularity Effects}

The results identified an additional explanatory factor is popularity, defined as the frequency with which terms and identifiers appear in biomedical text corpora (e.g., PubMed Central). Popularity reflects the likelihood that a fact will be encountered by the model during pretraining. Fine-tuning success was strongly correlated with identifier frequency: GO identifiers appeared more often than HPO identifiers, while gene symbols were orders of magnitude more frequent still (Table~\ref{tab:fact_popularity}).

This frequency effect parallels what recent theoretical work describes as \textit{factual salience}—the internal strength with which facts are encoded in model weights after pretraining~\cite{ghosal2024understanding}. In this sense, factual salience can be viewed as a manifestation of the popularity of a fact within the model’s parameters. Table~\ref{tab:fact_popularity} quantifies these patterns: GENE identifiers appeared approximately 200-fold more frequently in PMC than HPO identifiers, directly corresponding to the observed differences in fine-tuning success rates. Ghosal et al.~\cite{ghosal2024understanding} demonstrated that the number of times a fact is seen during pretraining correlates with its salience, implying that corpus-level popularity serves as a practical proxy for how strongly knowledge is stored. Highly salient facts—those with strong pretrained associations between a term and its identifier—are more easily retrieved and require fewer fine-tuning updates for consolidation. Conversely, low-salience facts, typically drawn from the long tail of rare identifiers, pass through a prolonged \textit{guessing phase} before stable memorization is achieved.

In a complementary view, Gekhman et al.~\cite{gekhman2024does,gekhman2025inside} propose that fine-tuning is most effective for partially known or latent facts that are already weakly represented in the model, whereas entirely unfamiliar facts are harder to integrate and may even promote hallucination. Similarly, Chang et al.~\cite{chang2024large} showed that popularity mediates the transition from memorization to generalization: as exposure increases, models shift from rote token-level learning to semantically grounded retrieval. Together, these perspectives suggest that popularity not only strengthens factual salience but also fosters the formation of semantic embeddings that enable generalization. Within this framework, popularity in text corpora can be viewed as an observable correlate of both factual salience and latent knowledge within the model’s representational space.

The distributional skew of this effect is evident in the Zipf-style log–log rank–frequency plot (Figure~\ref{fig:combined_freq_embed}~A), where gene symbols dominate the head of the distribution, GO identifiers occupy the mid-body, and HPO identifiers are concentrated in the long tail—consistent with the frequency gradient GENE~$>$~GO~$>$~HPO. This long-tail structure helps explain why fine-tuning yielded negligible gains for HPO: most sampled terms were rare, limiting the model’s ability to benefit from fine-tuning (Table~\ref{tab:derived_metrics}). This interpretation aligns with the view of Ghosal et al.~\cite{ghosal2024understanding}, who describe low-frequency facts as having low factual salience, and of Gekhman et al.~\cite{gekhman2024does,gekhman2025inside}, who argue that fine-tuning is most effective for partially known (latent) rather than entirely unfamiliar facts. Together, these perspectives situate popularity within a broader theoretical framework linking corpus frequency, knowledge salience, and fine-tuning efficiency.

Surprisingly, even though GO had lower PMC identifier counts than GENE, memorization results were greater for GO than for GENE. This finding aligns with our prior results~\cite{hier2025prior}, suggesting that fine-tuning is most effective in the \textit{reactive middle} of the term-frequency distribution. Fine-tuning is inefficient for very rare (tail) terms, which lack sufficient prior exposure for consolidation, and for very common (head) terms, which are already well encoded and exhibit ceiling effects~\cite{hier2025prior}. Because most GENE terms occupy the high-frequency end of the distribution (Figure~\ref{fig:combined_freq_embed}~A), they display limited additional gains from fine-tuning compared with GO terms.

Other studies (e.g., Wang et al.~\cite{wang2024re}) have shown that brute-force strategies (up to 100 training epochs) can improve performance on difficult-to-memorize HPO terms from the long tail, albeit inefficiently. Taken together, these results suggest that popularity—and its representational analog, factual salience—acts as a strong predictor of memorization success during fine-tuning until a ceiling is reached, where gains plateau due to extreme popularity. Nonetheless, popularity alone does not explain why the GENE model successfully mapped unseen term–identifier pairs from the validation set. To account for this, we next examine lexicalization.

\subsection*{Lexicalization Effects}

While popularity predicts the ease of memorization, only lexicalization explains why the GENE model generalized to unseen term–identifier pairs (Table~\ref{tab:all_terminology_performance},~\ref{tab:derived_metrics}). The outcome categories (Gainer, Loser, Correct, Incorrect) reveal that GENE achieved gains on both trained and validation terms, whereas GO gains were restricted to trained terms and HPO showed minimal gains overall (Table~\ref{tab:all_terminology_performance}, Figure~\ref{fig:all_sankey}). We use \textit{generalization} operationally to mean that the model correctly retrieved term–identifier pairs after fine-tuning that were not seen during training. The GENE model’s ability to map unseen protein–gene symbol pairs is best explained by the lexicalization of gene symbols, which provides semantic cues absent from the arbitrary codes of GO and HPO.

The term \textit{lexicalization} originated in traditional linguistics and refers to the process by which symbolic expressions or acronyms (e.g., radar, laser) become conventionalized as words in natural language~\cite{bennane2017lexicalization}. Even the term “LLM” is a contemporary example—it has undergone rapid lexicalization, evolving from an acronym to a widely recognized word in less than a decade. In the context of large language models, we extend this notion to describe how symbolic identifiers acquire semantic embeddings during pretraining that position them near meaningful words in vector space. We posit that an identifier is \textit{lexicalized} when its embedding aligns with that of its corresponding natural-language term, allowing the model to rely on semantic rather than purely sequential associations.

Our embedding analyses support this view. Cosine similarities based on Llama~3.1 embeddings (Figure~\ref{fig:semantic_alignment}~A) showed that protein–gene pairs exhibited significantly higher within-pair similarity than random comparisons, indicating above-chance semantic alignment. PCA visualizations (Figure~\ref{fig:combined_freq_embed}~B) reinforced this finding: gene symbols intermingled with protein names in embedding space, whereas HPO and GO identifiers remained clustered apart from their terms. Euclidean distance analyses (Figure~\ref{fig:semantic_alignment}~B) further confirmed that GENE term–identifier pairs were located much closer together than HPO or GO pairs.

Taken together, these results suggest that lexicalization provides a representational mechanism for generalization. When identifiers are arbitrary codes (HPO, GO), fine-tuning reinforces memorized token sequences. When identifiers are lexicalized (GENE), fine-tuning leverages semantic overlap to extend learning to unseen terms. Thus, lexicalization helps explain why generalization was observed exclusively in the fine-tuned GENE model. Our interpretation of generalization is closely aligned with recent work demonstrating that fine-tuning on even a small subset of memorized associations with semantically meaningful prompts can “unlock” broader latent knowledge already present in the model’s parameters~\cite{wu2025rote}. Once the model aligns a few examples with specific semantic cues, it can retrieve other facts encoded under the same relational structure — even if those specific instances were never presented during fine-tuning. This perspective matches our observation that correct predictions on a subset of protein–gene symbol pairs led to generalization across additional, unseen pairs, suggesting that fine-tuning acts less as a source of new knowledge and more as a mechanism for activating existing representations. In this sense, our results empirically support the theoretical framework of Wu et al. \cite{wu2025rote}, demonstrating that the same principle holds in the biomedical normalization domain, where lexicalized identifiers enable latent associations to be activated beyond the fine-tuned examples.

Importantly, these gains did not come without cost. As noted by Pletenev et~al.~\cite{pletenev2025much}, fine-tuning can degrade existing knowledge, and in our experiments, losses were greatest for the GENE model (Table~\ref{tab:derived_metrics}, column \textit{Degraded}). This dual outcome—generalization coupled with degradation—underscores the need to consider popularity and lexicalization as interacting influences during fine-tuning.

\subsection*{How Popularity Holds the Key to Lexicalization}

Popularity and lexicalization are best understood as interacting forces rather than independent explanations. At low levels of popularity, identifiers remain arbitrary, and fine-tuning yields little benefit (as with HPO). With moderate popularity (GO), repeated exposure strengthens token-level associations, allowing fine-tuning to support rote memorization. However, once popularity exceeds a critical threshold, a qualitative shift occurs: identifiers begin to acquire meaning through their co-occurrence patterns, and the model transitions from mere sequence memorization to embedding-based representation. This transition resembles a tipping point rather than a smooth continuum—an abrupt reorganization of the model’s representational space rather than a gradual progression.

Crucially, lexicalization is not created by fine-tuning but is inherited from pretraining. During pretraining, high-frequency identifiers repeatedly co-occur with their descriptive terms, allowing their embeddings to converge in semantic space \cite{periti2024lexical}. Fine-tuning then exploits this pre-established alignment by adjusting decision boundaries to make latent associations accessible. This interpretation aligns with Chang et al.~\cite{chang2024large}, who demonstrated that popular facts are not only more likely to be memorized but also more likely to develop semantically coherent embeddings that support generalization. In this view, popularity drives a representational phase change: repetition transforms statistical co-occurrence into stable, vectorized alignments that constitute lexicalization. Moreover, this tipping point or phase shift likely occurs during pretraining, not during fine-tuning.

Accordingly, rare mappings (unpopular facts) are difficult to memorize; moderately frequent mappings can be memorized but not generalized; and highly frequent mappings cross the lexicalization threshold, acquiring embeddings that support semantic alignment and generalization. Once this threshold is crossed, identifiers cease to be arbitrary tokens and function instead as meaningful lexical units. Generalization to unseen term–identifier pairs after fine-tuning thus emerges not from incremental exposure, but from a paradigmatic shift in how the model encodes meaning—one that reorganizes representation rather than merely reinforcing memorized sequences.

\subsection*{Limitations}

While our findings provide new insights into the mechanisms of fine-tuning success, we do acknowledge that there are some limitations. Note that we attributed all gains on training terms to memorization, although some may reflect embedding-based alignment. This yields a conservative estimate of generalization. We defined generalization as the ability of the fine-tuned model to learn unseen term–identifier pairs from the validation set, recognizing that the precise definition of generalization for factual fine-tuning remains unsettled. 
In addition, the balancing of test sets across head, body, and tail frequency distributions relied on PubMed Central identifier counts and ontology annotation counts as surrogates for pre-training exposure. Because true pre-training frequencies are unknown, these proxies are imperfect. The strong long-tail structure of HPO meant that most sampled terms were rare, partly explaining its low memorization rates (Table \ref{tab:derived_metrics}).

The mapping task itself—linking natural-language terms to arbitrary identifiers—represents a special case of fact memorization. This asymmetry applies to HPO and GO but not to protein–gene symbol pairs, and results may not generalize to symmetric fact pairs where both elements carry semantic content (e.g., Paris $\leftrightarrow$ France).
For this work, we analyzed embeddings only in the base Llama 3.1 8B model, assuming that fine-tuning re-weights token probabilities more than it reshapes embedding geometry. This assumption remains untested.
We also used a single PEFT setup (LoRA on Llama 3.1 8B) with fixed hyper-parameters. Alternative adapters, data-mixing strategies, or larger base models may alter the memorization–generalization trade-off. We also attributed knowledge loss (degradation) to parameter overwriting, though the mechanisms are likely related to catastrophic forgetting. 
Finally, although lexicalized identifiers appear to support generalization, the underlying mechanism remains uncertain. Whether fine-tuning primarily exploits pre-existing embedding structures or induces new representational dynamics is an open question. Moreover, this study does not address recent proposals that factual associations may be stored in localized \textit{knowledge neurons} or \textit{knowledge subnetworks} within transformer architectures~\cite{dai2021knowledge, niu2024does}.

\subsection*{Future Work}

Future studies should extend these findings in several ways.
First, larger and more comprehensive test sets are needed to confirm robustness. Expanding sampling across the full frequency range of ontology terms would strengthen evidence for the roles of popularity and lexicalization. Applying the same framework to additional terminologies, such as ICD-10, SNOMED CT, RxNorm and LOINC, would test whether the observed continuum from memorization to generalization generalizes across biomedical vocabularies.
In addition, evaluating larger base models (e.g., Llama 3.1 70B) and other foundation models (e.g., GPT-4o, Mistral, Claude) would clarify how scale and architecture influence the memorization–generalization balance. Systematic exploration of fine-tuning parameters—adapter types, learning-rate schedules, and data-mixing strategies—could reveal methods that reduce knowledge degradation while preserving generalization. Finally, direct analysis of embeddings in fine-tuned models may determine whether fine-tuning primarily re-weights existing representations or reshapes semantic structure.

\section*{Conclusion}

This study evaluated the capacity of large language models to normalize biomedical terms across three contrasting terminologies: the Human Phenotype Ontology (HPO), the Gene Ontology (GO), and protein–gene symbol mappings (GENE). Using Llama~3.1 (8B and 70B) and GPT-4o as base models, and fine-tuning the Llama~3.1~8B model, several consistent patterns emerged. The findings reveal that popularity—defined as exposure to term–identifier pairs during pretraining—proved essential for memorization. Large language models leverage popularity to consolidate \textit{factual salience}, the internal strength with which facts are encoded in model parameters. GO identifiers, which occur more frequently in biomedical text than HPO identifiers, supported robust memorization after fine-tuning, whereas the long-tailed, infrequent HPO identifiers showed only modest gains. In addition, \textit{lexicalization} emerged as the key to generalization. Protein names and gene symbols are semantically aligned, allowing the model to retrieve term–identifier pairs not seen during fine-tuning but already represented in model parameters through pretraining. By contrast, the arbitrary identifiers of GO and HPO constrained the model to rote memorization. Lexicalization appears to arise during pretraining, when highly frequent identifiers (high popularity) acquire meaningful embeddings (high factual salience).

Directionality and scale effects were consistent. Across all terminologies, mappings from \textit{term}$\rightarrow$\textit{identifier} uniformly outperformed \textit{identifier}$\rightarrow$\textit{term}, reflecting the autoregressive bias of large language models. Larger models consistently outperformed smaller ones, and fine-tuning the 8B model did not close the gap with Llama~70B or GPT-4o.
Knowledge gains came with trade-offs. Fine-tuning improved performance on both trained and some unseen terms but also degraded portions of preexisting knowledge—particularly for GENE mappings, where baseline accuracy was already high.

Together, these findings indicate that fine-tuning promotes memorization when identifiers have high popularity in the training corpus, whereas generalization emerges only when identifiers are lexicalized. This \textit{popularity–lexicalization continuum} provides a predictive framework for understanding when fine-tuning will succeed, when it will fail, and why some terminologies remain resistant to factual knowledge acquisition. Crossing the lexicalization threshold is not merely a gradual gain in memorization capacity,  but a  transition  from meaning encoded in token patterns to meaning residing in  semantically aligned structures.

\bibliography{sn-bibliography}
\raggedright
\section*{Author Information}

\subsection*{Author Affiliations}
\textbf{Suswitha Pericharla} \\[4pt] 
Computer Science Department, Missouri State University, Springfield, MO, USA.\\

\noindent  \textbf{Daniel B. Hier}\\
ORCID id: https://orcid.org/0000-0002-6179-0793 \\
Engineering Program, Missouri State University, Springfield, MO, USA.\\
Department of Neurology and Rehabilitation, University of Illinois, Chicago, IL, USA.\\[4pt]

\noindent \textbf{Tayo Obafemi-Ajayi}\\[4pt]
ORCID id: https://orcid.org/0000-0002-0155-9733\\
Engineering Program, Missouri State University, Springfield, MO, USA.

\subsection*{Author Contributions}
S.P., D.B.H., and T.O. conceived and designed the study. 
D.B.H. developed the data processing workflow while S.P. performed the experimental analysis. 
All authors contributed to writing and editing of the manuscript. T.O. supervised the project.

\subsection*{Corresponding Author}
Correspondence to Dr. Tayo Obafemi-Ajayi 
(\href{mailto:TayoObafemiAjayi@MissouriState.edu}{TayoObafemiAjayi@MissouriState.edu}).\\[8pt]

\section*{Declarations}

\subsection*{Ethics Approval and Consent to Participate}
This study did not involve human subjects or identifiable human data.

\subsection*{Consent for Publication}
All authors have approved the final manuscript for publication.

\subsection*{Availability of Data and Materials}
All data analyzed during this study are available on the Computational Learning Systems Lab github page: https://github.com/clslabMSU. Code used in the analysis is also available on the github page.

\subsection*{Competing Interests}
The authors declare that they have no competing interests.

\subsection*{Funding}
None.

\subsection*{Acknowledgements}
None.
\end{document}